\documentclass[letterpaper, 10 pt, conference]{ieeeconf}  
\usepackage{graphicx}

\IEEEoverridecommandlockouts                              

\usepackage{graphics} 
\usepackage{epsfig} 
\usepackage{mathptmx} 
\usepackage{times} 
\usepackage{amsmath} 
\usepackage{amssymb}  
\usepackage{xcolor}
\usepackage{cuted}
\usepackage{caption}
\usepackage{booktabs}

\definecolor{rblue}{rgb}{0,0.5,1}
\definecolor{awesome}{rgb}{1.0, 0.13, 0.32}
\definecolor{hollywoodcerise}{rgb}{0.96, 0.0, 0.63}
\definecolor{lasallegreen}{rgb}{0.03, 0.47, 0.19}
\definecolor{hanpurple}{rgb}{0.32, 0.09, 0.98}
\definecolor{green(pigment)}{rgb}{0.0, 0.65, 0.31}

\makeatletter
\let\NAT@parse\undefined
\makeatother
\usepackage[pagebackref=false, breaklinks=true, colorlinks, bookmarks=false]{
        hyperref
}
\hypersetup{
        colorlinks=true,
        linkcolor={red},
        citecolor={hanpurple},
        urlcolor={magenta}
}
\title{\LARGE \bf
RoboFind: Multi-Agent Personalized Object Search\\for People Who Are Blind or Have Low Vision
}
\author{Ruiping Liu$^{1,*}$, Shaofang Quan$^{1,*}$, Qian Yin$^{1}$, Jingqi Zhang$^{1}$, Junwei Zheng$^{1}$, Yufan Chen$^{1}$, Di Wen$^{1}$,\\Weijia Fan$^{1}$, Kailun Yang$^{2}$, M. Saquib Sarfraz$^{1}$, Tamim Asfour$^{1}$, Kunyu Peng$^{1,\dag}$, and Rainer Stiefelhagen$^{1}$  
\thanks{\dag\;Corresponding Author (email:~\href{mailto:kunyu.peng@kit.edu}{kunyu.peng@kit.edu})}
\thanks{*\; Co-first authors with equal contribution. 
(emails: \href{mailto:ruiping.liu@kit.edu}{ruiping.liu@kit.edu},
\mbox{\href{mailto:quanshaofang@gmail.com}{quanshaofang@gmail.com})}}
\thanks{\raggedright{$^{1}$The authors are with the Institute for Anthropomatics and Robotics, Karlsruhe Institute of Technology, Karlsruhe 76131, Germany.}
}
\thanks{\raggedright{$^{2}$The author is with the School of Artificial Intelligence and Robotics and the National Engineering Research Center of Robot Visual Perception and Control Technology, Hunan University, Changsha 410012, China.}}
}
\begin{document}

\maketitle
\thispagestyle{empty}
\pagestyle{empty}
\begin{strip}
\vskip-12ex
    \centering
    \includegraphics[width=\linewidth]{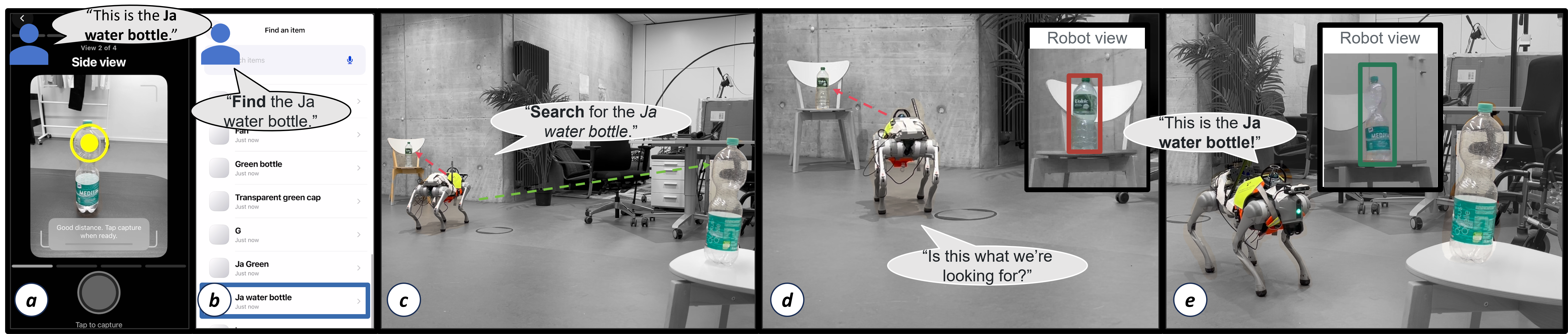}
    \vskip-1ex
    \captionof{figure}{Overview of RoboFind. (a) The user teaches a personal object once through guided smartphone recordings. (b) A later mission selects that object from the personal library. (c) The robot searches on its own. (d) At a stable stop, the candidate is verified against the stored references rather than accepted as the target. (e) Found is reported after verification succeeds.}
    \label{fig:teaser}
    \vspace{1em}
\end{strip}

\begin{abstract}
Blind and low-vision users often need to locate a specific personal object rather than an arbitrary instance of the same category. The task calls for a robot that can move through the space and reach viewpoints the user cannot, and for an accessible interface where the user says which object is meant and learns whether the right one was found. We present RoboFind, a multi-agent framework in which a smartphone teaches the target and a quadruped robot carries out the search. A Target Teaching Agent converts guided smartphone recordings into a semantic target profile and a reusable multi-view reference bank through an accessible capture flow with AR guidance, speech and haptic feedback, and screen-reader support, so later missions refer to a stored object without repeating the teaching process. At runtime, a Navigation Agent explores the environment and proposes candidate targets, a Verification Agent checks each candidate against the stored references, and a Coordination and Recovery Agent completes the mission or triggers recovery and continued search. Across 32 real-robot missions, RoboFind reaches 85.0\% success against 25.0\% for a reconstructed sequential first-stop baseline over 20 trials with ten targets, and reduces false success from 75.0\% to 5.0\%. On six shared targets it succeeds in 10/12 trials, against 5/12 for 12 independently executed GPT-6 Astra-only trials. These results show that the multi-agent design fits the demands of personalized object search, where verifying object identity before declaring completion is what makes the outcome something a user can rely on. The source code and results will be made publicly available.
\end{abstract}

\section{Introduction}

Robotic guide dogs have been investigated as mobility assistants for Blind and Low-Vision (BLV) people, supporting obstacle avoidance~\cite{liu2026not, hwang2023system}, route following~\cite{hwang2026guidenav}, and communication and coordination~\cite{hayamizu2026woofs, yu2026canine}. Everyday visual assistance, however, extends beyond moving between locations. BLV users often need to identify a particular possession, inspect its surroundings, or understand what is present on a table, without necessarily approaching or retrieving an object themselves~\cite{brady2013visual, gonzalez2024investigating}. User-operated smartphone and wearable systems~\cite{liu2026objectfinder, bemyeyes_smartglasses} place this burden on the user, who has to walk the space and orient the camera until the object falls into view. An object outside the field of view stays invisible to the recognizer, however well it has been trained. \emph{Can a robot search for a user's personal objects without requiring the user to accompany the search physically?} This question suggests a division of roles. The robot supplies the mobility and viewpoints that object search requires, rather than guiding the user's own movement, while the smartphone, already the most accessible interface for BLV users, is where the target is specified and the outcome is confirmed.

Personalization matters here because the target is often \emph{a particular object}, not any instance of its category. A request for \textit{``a black backpack''} can guide a category-level search, but leaves several similar objects as plausible matches. Prior work has explored how users can customize visual assistance to their individual needs~\cite{herskovitz2023hack, herskovitz2024programally}, while context-aware systems tailor visual descriptions to users' intents~\cite{worldscribe}. Find My Things~\cite{wen2025find} shows that users can teach a system to recognize their own belongings through an accessible interface. Teaching, however, only settles which object is meant. Whether the robot has actually reached that object is a separate question, and one the user cannot answer by looking. The challenge is therefore to connect personalized target specification, physical search, and instance-level verification.


To address this challenge, we introduce \textsc{RoboFind}, a multi-agent framework that decouples smartphone-based personal-object teaching from quadruped-executed search, illustrated in Fig.~\ref{fig:teaser}. The smartphone provides the interface for teaching and selecting targets, while the robot acquires observations and explores the environment. A \emph{Target Teaching Agent} converts guided smartphone recordings into a semantic target profile and a compact multi-view visual reference bank, which are stored in a reusable object library so that later missions refer to an existing target without repeating the teaching process.

At runtime, a \emph{Navigation Agent} uses the stored target description to navigate and propose stopping candidates.
A \emph{Verification Agent} checks each candidate against the stored target references, rather than accepting the navigation proposal by default. A \emph{Coordination and Recovery Agent} completes the mission after PASS or triggers recovery and continued search after REJECT or UNCERTAIN. This feedback closes the loop between search and verification.

We evaluate the framework in 32 physical missions. RoboFind is tested in 20 trials over ten targets, and GPT-6 Astra-only~\cite{openai2026gpt6astra} is tested in 12 independent trials over six of these targets. The sequential first-stop baseline is reconstructed from the same 20 RoboFind runs by terminating each mission at its first stable candidate, which represents how a VLN policy alone would handle this task and yields a trial-level paired comparison over identical scenes and starting conditions. RoboFind succeeds in 17/20 trials, compared with 5/20 for the sequential baseline. On the six shared targets, RoboFind succeeds in 10/12 trials, compared with 5/12 for GPT-6 Astra-only. Beyond success rates, these experiments also examine robot-side task completion, runtime cost, and corner cases. 
We further envision potential assistive tasks after the target has been located and verified, outlining how RoboFind's multi-agent framework could support accessible scene understanding and object interaction beyond search.
Our contributions are threefold:
\begin{itemize}
\item We develop a personalized robotic object-search system in which a smartphone defines and confirms the target through an accessible teaching interface and a reusable object library, while a quadruped robot carries out the physical search.

\item We design a multi-agent framework with four specialized roles for teaching, navigation, verification, and recovery, in which verification feedback closes the loop by returning an unaccepted candidate to continued search.

\item We validate the framework on a real robot against a paired sequential reconstruction and an MLLM-only baseline, showing that the multi-agent design fits the demands of personalized object search.

\end{itemize}


\section{Related Work}

\subsection{Vision-Language Navigation for Robot Guide Dogs}

Robot guide dogs require embodied navigation systems that combine environmental perception, instruction following, user safety, and accessible communication~\cite{hwang2023system,cai2024navigating}. User studies and co-design further identify interaction preferences and practical deployment requirements~\cite{doore2024co_designing,kim2025understanding, biggs2026terrain}. 
Complementing robot-level navigation, ground-surface recognition and tactile walking surface indicator perception support awareness of terrain and accessibility infrastructure~\cite{bazhenov2024dogsurf,hwang2026guidetwsi}. Prior work also explores user-informed visual navigation~\cite{hwang2026guidenav}, verbal communication~\cite{hayamizu2026woofs}, and interactive user coaching~\cite{yu2026canine}. General-purpose vision-language navigation~\cite{zhang2025uninavid,ding2026uni}, personalized instance navigation~\cite{barsellotti2024personalized}, and human-aware hazard perception~\cite{liu2026not} provide complementary foundations for assistive navigation that considers both user-specific targets and the different perceptual requirements of the robot and its user.

\subsection{Multi-Agent Collaboration in Embodied Systems}

Multi-agent embodied systems investigate task decomposition, cooperative planning, and execution across heterogeneous agents~\cite{kang2025viki,qin2025robofactory, zu2025collaborative_tree_search}.
MICA coordinates role-specialized agents for industrial assistance~\cite{wen2025mica}, while ComBodied Agents proposes a broader human-centric paradigm for agentic AI~\cite{ding2026combodied}. Collaborative multimodal reasoning has also been explored in long-video understanding and reference audio-visual segmentation, where agents exchange information and refine predictions through iterative collaboration~\cite{chen2025lvagent,zhao2026mar3}. More directly, GuideFetch coordinates concurrent navigation and object retrieval for assistive robot dogs~\cite{yin2026guidefetch}, while Seeing Together studies cooperative spatial reasoning from multiple robots' egocentric observations~\cite{peng2026seeing}. These works motivate role-specialized collaboration in embodied systems. Our work extends this idea to personalized object search: smartphone-based teaching defines the target, robot navigation proposes candidates, and instance verification determines whether the search terminates or continues.

\section{Methodology}
\subsection{Procedure Overview}
\label{subsec:procedure_overview}

RoboFind supports personalized object finding for blind and low-vision users through two stages: \textbf{target teaching} and \textbf{autonomous search}. During teaching, the user names an object and records four guided videos on the smartphone. The server processes these recordings into a semantic target profile, which carries the object category and the navigation instruction, and a compact multi-view visual reference bank. Both representations are stored and reused across subsequent search missions, avoiding repeated teaching of the same object.

To initiate a search, the user selects a registered object in the application. A vision-language-navigation model searches for and approaches objects according to the stored navigation instruction. A stable stopping event produces a candidate for instance verification against the stored references. An accepted candidate completes the mission, while a not verified one triggers recovery and search resumption. This closed loop separates navigation-level candidate discovery from instance-level acceptance.

The system comprises three components: a smartphone for teaching and mission initiation, a cloud server hosting the models and mission services, and a Unitree Go2 robot providing egocentric RGB observations and executing motion commands.

\begin{figure}
    \centering
    \includegraphics[width=\linewidth]{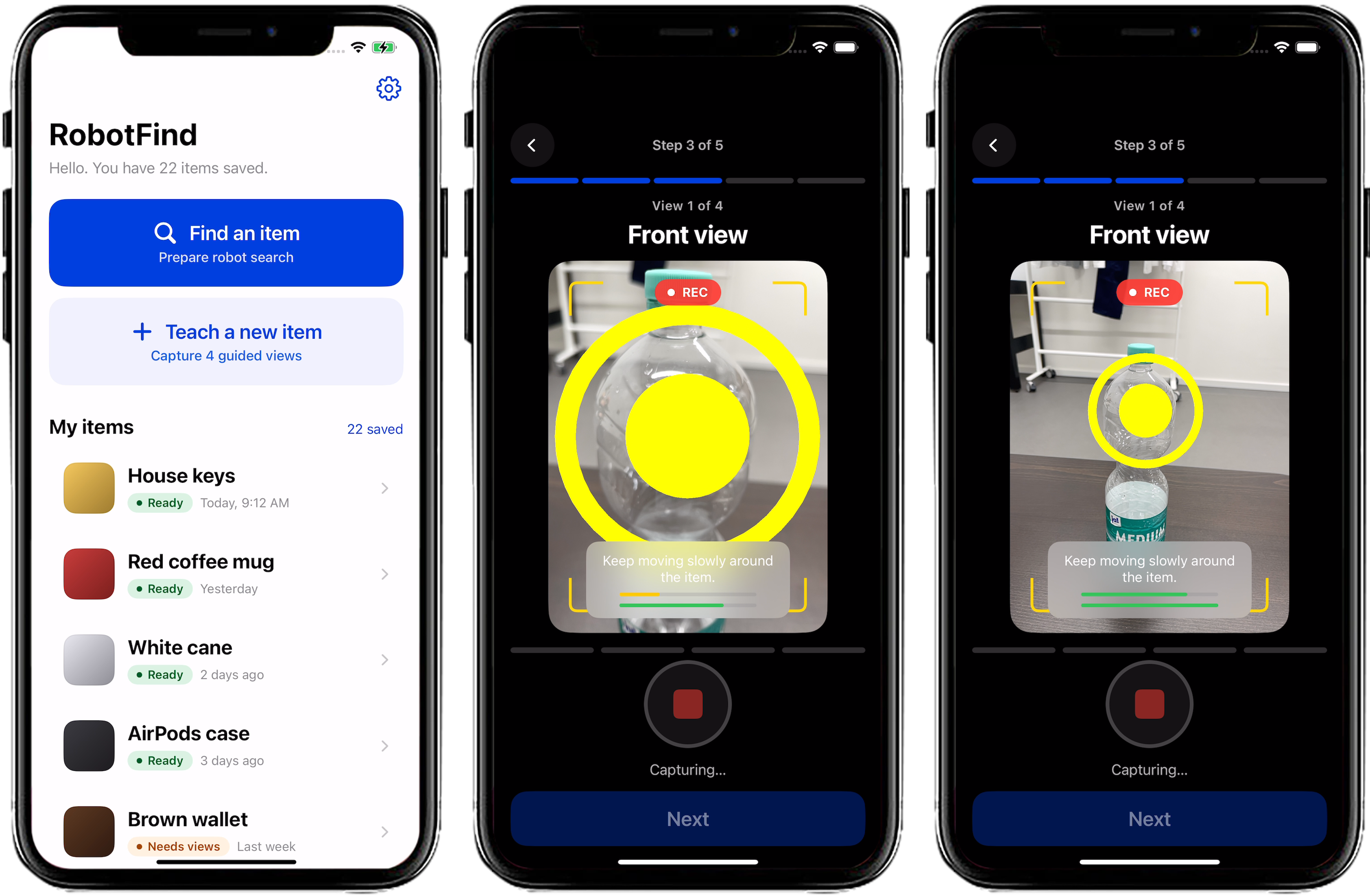}
    \vskip-1ex
    \caption{RoboFind smartphone interface for object teaching and search. The yellow marker provides AR-based capture guidance.}
    \label{fig:UI}
    \vskip-3ex
\end{figure}

\begin{figure*}
    \centering
    \includegraphics[width=0.95\linewidth]{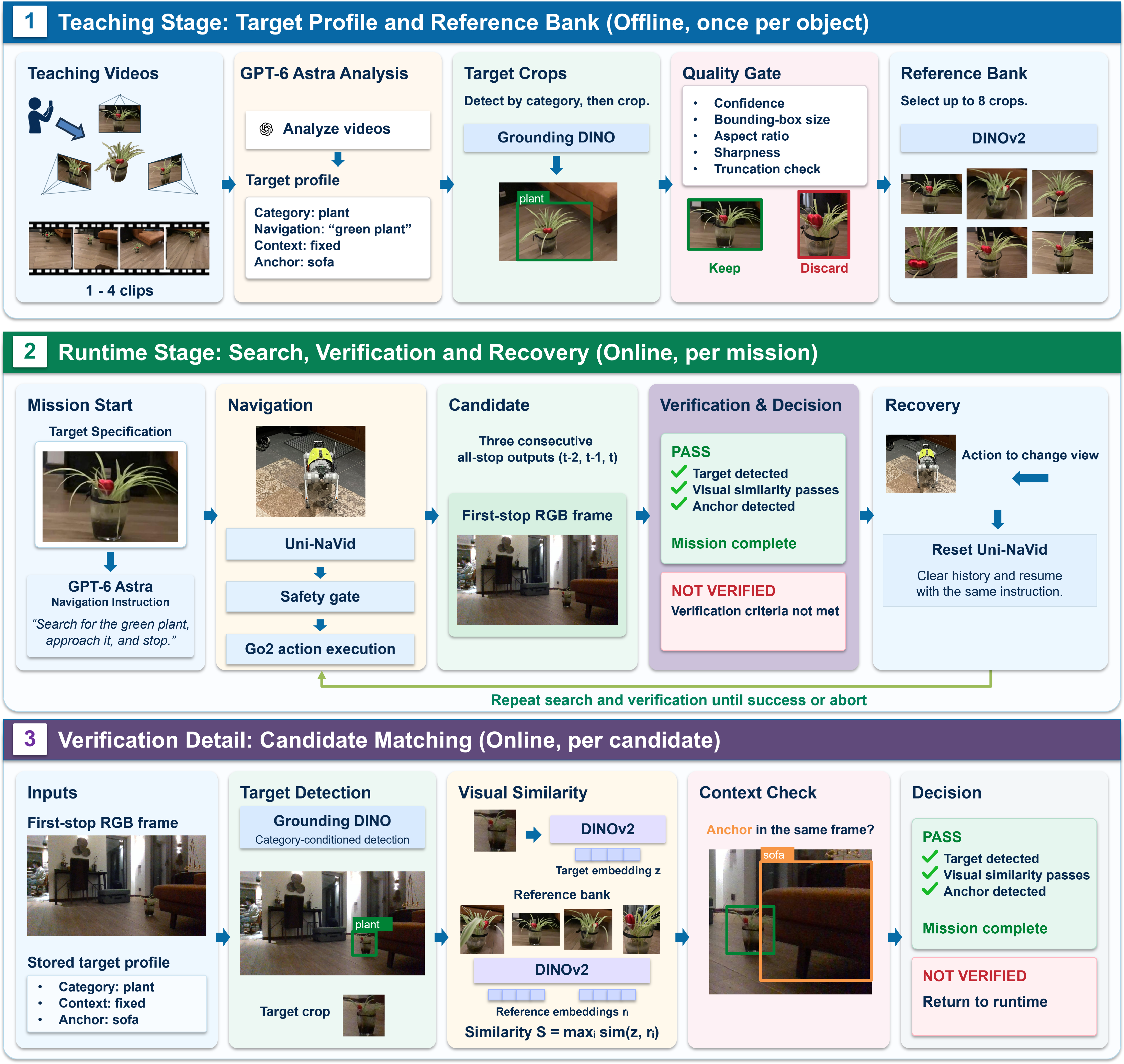}
    \caption{RoboFind multi-agent pipeline for target teaching, robot search, candidate verification, and recovery.}
    \label{fig:pipeline}
    \vskip-3ex
\end{figure*}
\subsection{User Interface Design}
\label{subsec:user_interface_design}

The RoboFind user interface is implemented in SwiftUI and organized around two primary actions: \textbf{Teach a new item} and \textbf{Find an item}, as shown in the leftmost image in Fig.~\ref{fig:UI}. A local item library provides access to previously registered objects. Large controls, explicit text labels, high-contrast visual elements, speech guidance, haptic feedback, and VoiceOver support accessible interaction and reduce reliance on visual input. The user interface and accessible interaction design follow the established interaction paradigm of Find My Things~\cite{wen2025find}.

\paragraph{Guided teaching.}
Teaching asks two things of the user, naming the item and recording the guided views. The app handles the rest, walking through capture preparation, analyzing the recordings, and presenting the resulting profile for review. The four recordings cover the front, side, top, and an additional view with a different background. During each recording, an AR reference point provides movement guidance, as shown in the last two images in Fig.~\ref{fig:UI}. This point represents the initial capture position rather than a detected object. A clip is completed when the phone has moved approximately $0.40$\,m relative to the initial position and the reference point has remained within the camera viewport for a cumulative duration of at least four seconds. Speech and haptic feedback supplement the visual progress indicators.



\paragraph{Item management and search feedback.}
After teaching, users can review the generated profile, save the item to the personal library, or run a trial search to check that the object is recognized. During search, the interface exposes only four high-level states: \textit{Searching}, \textit{Found}, \textit{Stopped}, and \textit{Search failed}. Navigation, verification, and recovery remain hidden from the user. \textit{Found} is reported only after the target has been verified, rather than when the robot merely stops.

\subsection{Multi-Agent Framework}
\label{subsec:multi_agent_framework}

We organize the system into four specialized roles, shown in Fig.~\ref{fig:pipeline}. A \textbf{Target Teaching Agent} builds the target representation, a \textbf{Navigation Agent} discovers candidates, a \textbf{Verification Agent} decides acceptance, and a \textbf{Coordination and Recovery Agent} controls the runtime.

\subsubsection{Target Teaching Agent}
\label{subsubsec:target_teaching_agent}

The Target Teaching Agent converts smartphone recordings into reusable semantic and visual representations. GPT-6 Astra analyzes the videos to generate a structured target profile
\begin{equation}
P = \{c, d_{\mathrm{nav}}, m, a\},
\end{equation}
where $c$ is the object category, $d_{\mathrm{nav}}$ is the navigation instruction, $m$ indicates whether the object is movable or context-stable, and $a$ is an optional contextual anchor. For example, a personal cup is movable and can be identified by its appearance alone. In contrast, an office chair at a specific workstation is context-stable, since the same office may contain many similar chairs, and it is therefore identified by its location context. 
The navigation instruction contains only the category and, optionally, a primary color, such as ``black umbrella.'' This keeps the instruction within the phrasing that object-goal navigation models are trained and evaluated on, where the target is given as a category label with a few attributes~\cite{batra2020objectnav, yokoyama2024hm3d}. 
Richer personal detail would fall outside what the navigator can ground, so adding it would put identity decisions in a module that has no way to check them. Instance identity is therefore left entirely to the Verification Agent.

The agent also constructs a multi-view reference bank. Grounding DINO~\cite{liu2024grounding} localizes the target in sampled teaching frames. Candidate crops are filtered using detection confidence, object size, sharpness, and boundary clipping, then encoded into L2-normalized, 768-dimensional DINOv2 embeddings~\cite{oquab2023dinov2}. Quality- and diversity-based selection retains at most eight reference crops, providing a compact visual memory across teaching viewpoints. 

\subsubsection{Navigation Agent}
\label{subsubsec:navigation_agent}

The Navigation Agent uses Uni-NaVid~\cite{zhang2025uninavid} to search for and approach potential targets. It receives egocentric RGB observations and a stored navigation instruction, such as \textit{``Search for a black umbrella, move close to it, and stop.''} Its high-level actions include forward, left, right, and stop, which are translated into Go2 velocity commands~\cite{anderson2021sim} and passed through a safety gate that suppresses commands violating a minimum clearance.

A single stop prediction does not establish mission success. A candidate is generated only after \textbf{three consecutive all-stop inference outputs}. The robot is then stopped, the current navigation session is terminated, and the stored first-stop RGB image is submitted for verification. 

\subsubsection{Verification Agent}
\label{subsubsec:verification_agent}

The Verification Agent determines whether a stopping candidate should be accepted as the taught target. Grounding DINO first localizes the requested category in the candidate image. If several detections are available, the highest-confidence detection is selected. Its crop is encoded using the same DINOv2 encoder employed during teaching and compared with the reference bank:

\begin{equation}
S
=
\max_{1 \leq i \leq N}
\hat{e}_c^\top \hat{e}_i,
\end{equation}

where $N \leq 8$ is the number of stored reference images, $\hat{e}_c$ is the normalized embedding of the candidate, and $\hat{e}_i$ is the normalized embedding of the $i$-th reference image. Appearance verification passes only if the requested category is detected and $S \geq T_{\mathrm{app}}$, where $T_{\mathrm{app}}$ is set empirically to 0.6 in our deployment. For movable targets, a detected candidate with $S < T_{\mathrm{app}}$ is marked as \textit{UNCERTAIN}, while a missing target detection is \textit{REJECTED}. For context-stable targets, acceptance additionally requires the stored anchor to be detected in the same image after appearance verification succeeds. This checks \textbf{anchor co-presence}, rather than explicit spatial relations such as ``beside'' or ``under.'' Candidates that fail appearance or context verification are \textit{REJECTED}. Both \textit{UNCERTAIN} and \textit{REJECT} are treated as NOT VERIFIED and trigger the same recovery procedure.

\subsubsection{Coordination and Recovery Agent}
\label{subsubsec:coordination_recovery_agent}


The Coordination and Recovery Agent is a deterministic runtime controller that manages transitions between search, verification, recovery, and termination. During verification, the robot remains stationary and the verifier does not issue motion commands. An accepted candidate completes the mission. Otherwise, the controller executes a predefined action, clears the previous Uni-NaVid history, and restarts navigation with the same navigation instruction. This fixed recovery cycle continues until a candidate is accepted, a non-recoverable error occurs, or the mission is terminated.

\section{Experiments}
\begin{table*}[t]
\centering
\small
\setlength{\tabcolsep}{4pt}
\caption{Ten-target comparison, with two trials per target.
$T_{\mathrm{succ}}$ averages successful trials only. Manifest outcomes include one interrupted RoboFind run labeled timeout.}
\label{tab:system_results_all10}
\vskip-1ex
\begin{tabular}{lrrrrr}
\toprule
Method & Trials & Success & False success & Timeout & $T_{\mathrm{succ}}$ (s) \\
\midrule
Sequential first-stop & 20 & 5/20 (25.0\%) & 15/20 (75.0\%) & 0/20 (0.0\%) & 160.3 \\
RoboFind (Ours) & 20 & \textbf{17/20 (85.0\%)} & \textbf{1/20 (5.0\%)} & 2/20 (10.0\%) & 225.5 \\
\bottomrule
\end{tabular}
\vskip-3ex
\end{table*}

\subsection{Experimental Setup}
\label{subsec:experimental_setup}

We evaluate RoboFind, our multi-agent search-and-verification framework, on a real Unitree Go2 robot with ten target objects and two trials per target. The evaluation covers household scenes, including a bedroom, living room, bathroom, kitchen, entrance hall, and garden. Six targets are configured as context-stable and four as movable, and depending on the target, the environment contains zero to two designated distractors. RoboFind combines phone-based teaching, Uni-NaVid navigation, candidate verification, and recovery. All teaching recordings were captured following the Find My Things capture tutorial~\cite{wen2025find}. When navigation comes to a stable stop, the stop yields a candidate for verification rather than an immediate decision that the target has been found. An accepted candidate completes the mission, whereas a NOT VERIFIED decision triggers recovery and continued search. Here, \emph{closed-loop} refers to this task-level verification feedback.

RoboFind is evaluated in 20 physical trials, and GPT-6 Astra-only is evaluated in 12 additional physical trials covering six of the ten targets, giving 32 executed missions in total. The sequential first-stop baseline is reconstructed from the RoboFind runs and requires no separate execution, as described in Sec.~\ref{subsec:baselines}. Counting each method separately, the evaluation comprises 52 method-level trial records. 
Tab.~\ref{tab:system_results_all10} reports the full ten-target comparison between RoboFind and the sequential baseline, and Tab.~\ref{tab:system_results_shared6} reports a separate three-method comparison on the six shared targets. The RoboFind and sequential results in the latter table are subsets of those in the former, not additional runs.

For each mission, we log navigation, verification, recovery, and termination events, together with the robot trajectory. Mission outcomes are manually annotated independently of the automatic verifier. Candidate stops, on the contrary, are labeled with an experimenter-defined observation rule. 

\subsection{Baselines}
\label{subsec:baselines}

\paragraph{Sequential baseline} 
Sequential denotes a one-way teaching-to-navigation pipeline without candidate-level verification feedback, reflecting how a VLN policy would be applied to personal-object navigation on its own. The target is given as a category label, optionally with attributes~\cite{batra2020objectnav, yokoyama2024hm3d}, and a stable stop is taken as task completion, following the standard convention in that literature~\cite{batra2020objectnav}. The baseline therefore measures what the navigation model alone delivers, rather than a deliberately weakened variant. Since such a pipeline is a prefix of our closed-loop execution, we reconstruct it from the same 20 RoboFind runs by taking the outcome at the first stable candidate, which keeps the scene, the taught references, the navigation policy, and the starting pose identical across conditions. One consequence is that the reconstructed baseline cannot time out, since every reconstructed trial ends at its first candidate.

\paragraph{GPT-6 Astra-only}
We evaluate independently executed GPT-6 Astra navigation runs at their first recorded stable stopping candidate. Two trials are conducted for each of six targets: suitcase, green plant, backpack, trash can, towel, and umbrella. 

\subsection{Evaluation Metrics}
\label{subsec:evaluation_metrics}

\paragraph{Task outcomes}
We report success, false-success, and timeout rates over all evaluated trials. Success requires correct-target completion according to the manual annotation. False success denotes an incorrect terminal target, while timeout is failure to complete within the 600~s protocol limit.

\paragraph{Time and trajectory}
Time to target is measured from navigation start to successful completion and is reported only for successful trials. Trajectory metrics are computed from recorded odometry and include path length, net displacement, and accumulated yaw.

\paragraph{Verification and recovery cost}
We report the number of candidates and recoveries, client-side verification latency, and recovery duration. Verification latency includes communication and polling. We additionally report a load-adjusted value obtained by subtracting an estimated 20.6~s cold-start cost from two requests. Mission completion times are left unchanged.

\paragraph{Stage-wise runtime}
For successful missions, completion time is decomposed into initial search, verification, recovery, and residual post-candidate execution time. The residual includes subsequent search and other unallocated runtime.
\subsection{Experimental Results}
\label{subsec:experimental_results}

\paragraph{RoboFind versus sequential execution}
Tab.~\ref{tab:system_results_all10} reports the complete ten-target evaluation. RoboFind succeeds in 17/20 trials (85.0\%), with one false success and two timeouts. The sequential baseline succeeds in 5/20 trials (25.0\%) and has 15 false successes. Relative to terminating at the first stable candidate, the complete framework increases the observed success rate by 60.0 percentage points and reduces the false-success rate from 75.0\% to 5.0\%. Verifying each candidate before termination therefore turns most incorrect stops into correct completions. Instance-level verification combined with continued search is thus central to reliable personal-object search.


\paragraph{Three-method comparison on shared targets}
Tab.~\ref{tab:system_results_shared6} compares all three methods on the common target subset. Our framework succeeds in 10/12 trials (83.3\%), compared with 5/12 (41.7\%) for GPT-6 Astra-only and 4/12 (33.3\%) for the sequential first-stop baseline. No false success is observed for our framework in this subset, compared with seven for Astra and eight for the sequential baseline, although our framework receives two timeout outcomes.

\begin{table*}[t]
\centering
\small
\setlength{\tabcolsep}{4pt}
\caption{Three-method comparison on six shared targets, with two trials per target.}
\label{tab:system_results_shared6}
\vskip-1ex
\begin{tabular}{lrrrrr}
\toprule
Method & Trials & Success & False success & Timeout & $T_{\mathrm{succ}}$ (s) \\
\midrule
Sequential first-stop & 12 & 4/12 (33.3\%) & 8/12 (66.7\%) & 0/12 (0.0\%) & 177.1 \\
GPT-6 Astra-only~\cite{openai2026gpt6astra} & 12 & 5/12 (41.7\%) & 7/12 (58.3\%) & 0/12 (0.0\%) & 134.6 \\
RoboFind (Ours) & 12 & \textbf{10/12 (83.3\%)} & \textbf{0/12 (0.0\%)} & 2/12 (16.7\%) & 225.4 \\
\bottomrule
\end{tabular}
\vskip-3ex
\end{table*}

\begin{figure}[!t]
    \centering
    \includegraphics[width=\linewidth]{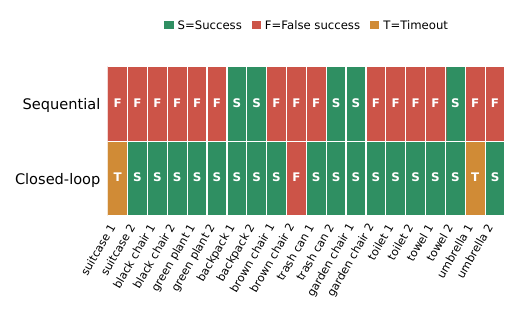}
    \vskip-1ex
    \caption{Paired outcomes for the reconstructed sequential first-stop baseline and RoboFind (shown as Closed-loop) across 20 trials.}
    \label{fig:paired_outcomes_detailed_landscape}
    \vskip-3ex
\end{figure}

\paragraph{Continuation beyond the first candidate}
The paired ten-target analysis (Fig.~\ref{fig:paired_outcomes_detailed_landscape}) shows where the additional successful trials arise. All five trials with a correctly annotated first candidate remain successful. Of the 15 trials with an incorrectly annotated first candidate, 12 ultimately succeed, one ends in false success, and two receive timeout outcomes. 
Thus, 12/15 initially incorrect candidates are followed by successful completion when execution continues through our framework. Overall, 12/17 successful trials (70.6\%) accept the target after the first candidate. Most of the gain comes from trials that a sequential first-stop pipeline would have ended at the wrong object. Continued search recovers these trials without affecting those that were already correct.

\paragraph{Candidate decisions and recovery}
Across the 20 RoboFind trials, the verifier processes 48 candidates. Under the recorded annotation protocol, it accepts 17 correct candidates and one incorrect candidate, while three correct and 27 incorrect candidates are not accepted. The 30 NOT VERIFIED decisions comprise eight UNCERTAIN and 22 REJECT outcomes, all followed by recovery. Recovery occurs in 14 trials, of which 12 eventually succeed. The framework processes an average of 2.4 candidates and executes 1.5 recoveries per trial. The verifier thus filters out nearly all incorrect stops, and recovery turns these rejections into renewed and mostly successful search.

\begin{figure}[t]
    \centering
    \includegraphics[width=0.9\columnwidth]{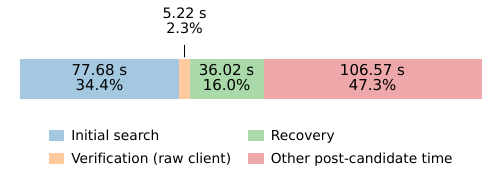}
    \vskip-1ex
    \caption{Mean cumulative runtime across 17 successful RoboFind missions, using pooled completion time. Segments aggregate repeated events rather than forming a chronological trace.}
    \label{fig:stagewise_runtime}
    \vskip-3ex
\end{figure}

\paragraph{Stage-wise runtime analysis}
Fig.~\ref{fig:stagewise_runtime} decomposes the mean completion time of 225.5~s across the 17 successful missions. Initial search accounts for 77.7~s (34.4\%), raw client-side verification for 5.2~s (2.3\%), and logged recovery for 36.0~s (16.0\%). The remaining 106.6~s (47.3\%) covers subsequent search and other unallocated execution time, rather than a directly measured navigation phase. Both cold starts are retained in this accounting. Across all 48 candidates, mean raw client verification latency is 1.9~s, or 1.1~s after the stated loading adjustment, and the adjusted verification total averages 2.6~s per trial. Verification therefore represents only a small fraction of elapsed execution, whereas recovery and other post-candidate execution account for a substantial share.

\paragraph{Trajectory analysis}
The reconstructed sequential first-stop baseline travels 2.4~m on average, which is by construction the leading portion of the corresponding RoboFind trajectory, compared with 2.7~m for GPT-6 Astra and 7.4~m for RoboFind using mission-local trajectory recomputation. The historical summary-path mean for RoboFind is 10.4~m across the same 20 trials. RoboFind also accumulates more rotation, reflecting the additional motion introduced by recovery and continued search. A logged recovery event lasts 22.6~s and adds only 0.1~m of translation but approximately $51.0^\circ$ of yaw on average. The additional path length thus comes from continued search, while recovery itself remains compact and gives navigation a new starting perspective at little translational cost.

\begin{figure*}
    \centering
    \includegraphics[width=0.9\linewidth]{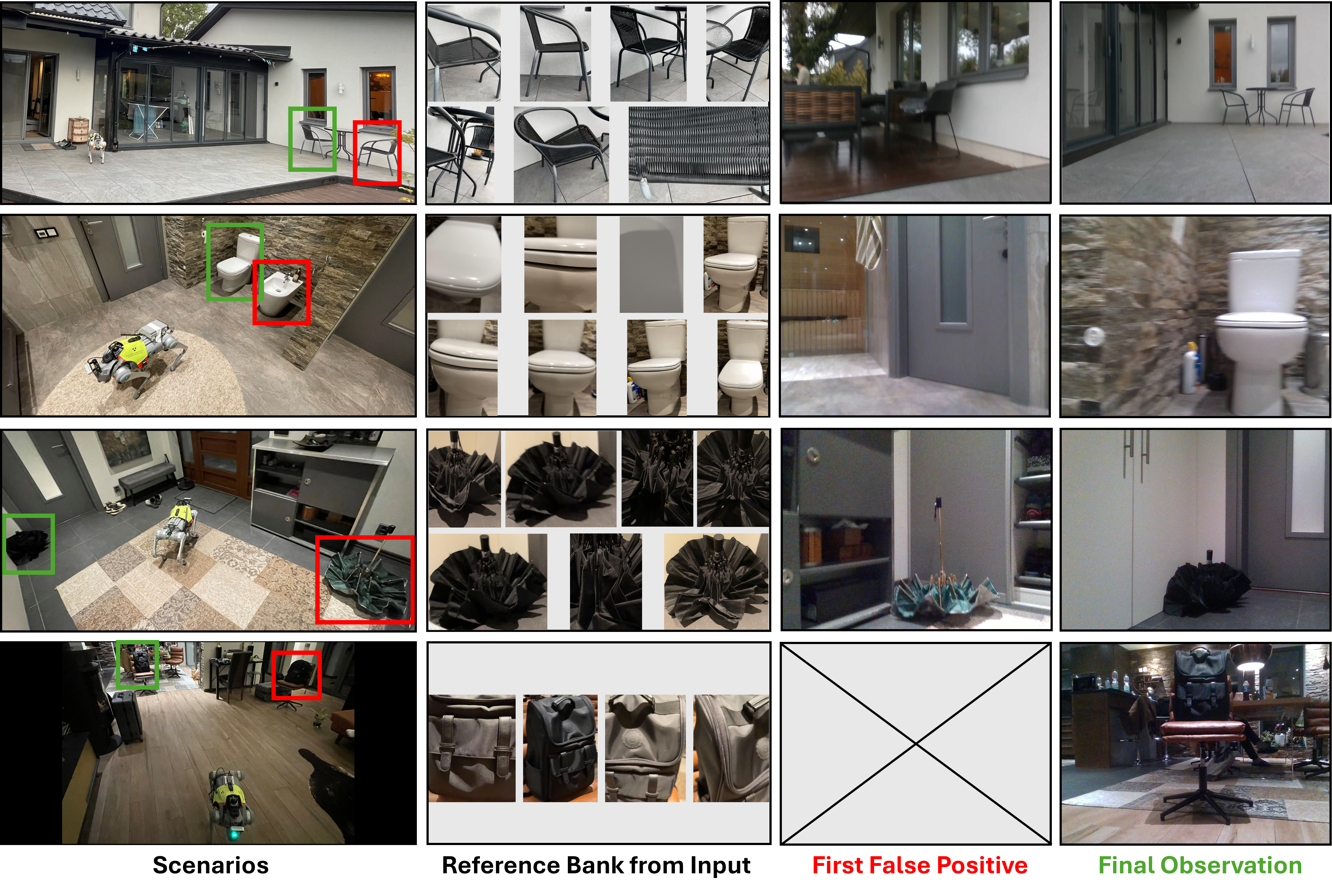}
    \vskip-1ex
    \caption{Qualitative examples showing scenarios, reference banks, false-positive candidates, and final observations.}
    \label{fig:qualitative_examples}
    \vskip-3ex
\end{figure*}
\paragraph{Execution time}
The median time to target across the 17 successful RoboFind trials is 165.8~s. On the six shared targets, mean successful-trial times are 225.4~s for our framework, 134.6~s for Astra, and 177.1~s for the sequential first-stop baseline. These means use different successful-trial subsets and therefore do not establish a matched-case speed ranking. RoboFind thus trades a moderate increase in completion time for a substantially higher rate of correct-target completion. 

\subsection{Qualitative Results}
Fig.~\ref{fig:qualitative_examples} shows representative missions for four targets. Each scene contains the taught target (green box) together with a same-category distractor (red box). For the garden chair, toilet, and umbrella, navigation first stops at an incorrect candidate, which the verifier does not accept because it does not match the stored reference bank. After recovery, the robot resumes search and reaches the taught object, as shown in the final observations. For the backpack, the first stop already reaches the correct target and is accepted directly. These examples illustrate how verification against personal references separates the taught object from visually similar alternatives.
\subsection{Corner-Case Analysis}
\label{subsec:corner_cases}

\paragraph{GPT-6 Astra-only}
The baseline exhibits both incorrect stopping and non-completing search. Six trials produce incorrect stable candidates despite three consecutive STOP outputs, indicating that stopping consistency does not establish personal-object identity. In a separate green-plant trial, 62 of 100 navigation records contain only turning actions, and the run is stopped without a stable candidate. Consecutive stop outputs thus reflect navigation confidence but not object identity, which an MLLM navigator alone cannot guarantee.

\paragraph{Our multi-agent framework}
Our framework treats navigation stops as candidate proposals, but verification can also fail. The single false-success trial accepts an incorrect brown-chair candidate with $S=0.672$ and a context MATCH. In the suitcase timeout, a correct candidate receives UNCERTAIN at $S=0.507$. Conversely, a garden-chair trial rejects a correct third candidate at $S=0.458$ but subsequently succeeds at candidate five, showing that an intermediate true-target non-acceptance need not determine the final outcome. All 12 executed context checks return MATCH, including the false acceptance, so these observations do not establish an additional rejection benefit from context checking. The other timeout, for umbrella, records a mission-API interruption at 312.5~s. The remaining failures arise from borderline similarity scores and runtime interruptions, pointing to adaptive thresholds and more robust mission services as promising next steps.


\section{Discussion}
\label{sec:discussion}

\paragraph{Delegated search that users can trust}
Find My Things demonstrates teachable object recognition for blind and low-vision users~\cite{wen2025find}. RoboFind extends this paradigm by letting users define \emph{what to find} once through a reusable personal-object library, while the robot takes over the physical effort of exploring a room and acquiring observations. Delegation, however, is only useful if users can rely on the outcome. Blind and low-vision users often cannot visually confirm whether the robot has reached the right object, so a false success may send them to the wrong item or leave them believing the search is finished~\cite{macleod2017understanding, alharbi2024misfitting}. For this reason, RoboFind reports \emph{Found} only after the target has been verified against the references taught by the user. Future interfaces could go further by conveying the robot's confidence, describing the accepted object so that users can judge it themselves, and asking for confirmation when a candidate remains ambiguous. Feedback during teaching could likewise help users record views that make their objects easier to verify later.

\paragraph{From finding to understanding and acting}
Locating an object is often only the first step toward a user's actual goal. Once the target is verified, RoboFind could describe its location and surroundings or answer questions such as \textit{``What is on the table?''}, building on intent-aware description systems such as WorldScribe~\cite{worldscribe}. With an arm, the same pipeline could extend to switching on a fan, bringing a personal remote control, or fetching a cup of coffee~\cite{yin2026guidefetch}. Such extensions would require spatial feedback that is meaningful to the user, planning grounded in executable robot skills~\cite{ahn2022saycan}, and verification of each action so that users know when a task has truly been completed.

\paragraph{Preserving user choice}
Delegating search to a robot should expand the options available to users rather than replace their own agency. ShelfHelp~\cite{agrawal2024shelfhelp} supports users in retrieving objects themselves, and Beyond Omakase~\cite{kamikubo2025beyond} highlights the value of shared control in assistive navigation. Future interfaces could let users decide, for each request, whether to receive information, act themselves, or hand the task to the robot, with accessible progress updates and the option to interrupt or redirect the search. Co-design sessions and user studies with blind and low-vision participants will be essential to understand how delegated search fits into everyday routines and which levels of assistance they prefer.

\section{Conclusion}
\label{sec:conclusion}

We presented \textsc{RoboFind}, a multi-agent framework connecting smartphone-based personal-object teaching with quadruped-executed search, in which verification feedback rather than navigation confidence determines whether the mission completes or the search continues. Real-robot experiments show higher observed success rates than a paired sequential first-stop reconstruction and an independently executed Astra baseline on shared targets. The results highlight the effectiveness of deciding termination by instance identity, while the corner-case analysis and discussion indicate future directions for the multi-agent framework and the surrounding system.

\noindent\textbf{Generative AI use disclosure.}
OpenAI Codex assisted in drafting and debugging parts of the experimental code and in editing the manuscript for clarity and language. The authors independently reviewed and validated all AI-assisted code and text. Every reported result was derived from actual experimental runs and verified by the authors; no empirical values were generated, modified, or fabricated by AI.

\section*{Acknowledgment}
The project is funded by the Deutsche Forschungsgemeinschaft (DFG, German Research Foundation) – SFB-1574 – 471687386. This work was supported in part by the SmartAge project sponsored by the Carl Zeiss Stiftung (P2019-01-003; 2021-2026). The authors gratefully acknowledge the computing time provided on the high-performance computer HoreKa by the National High-Performance Computing Center at KIT (NHR@KIT). This center is jointly supported by the Federal Ministry of Education and Research and the Ministry of Science, Research and the Arts of Baden-Württemberg, as part of the National High-Performance Computing (NHR) joint funding program (https://www.nhr-verein.de/en/our-partners). HoreKa is partly funded by the German Research Foundation (DFG).

\bibliographystyle{IEEEtran}
\bibliography{main}

\end{document}